\pdfoutput=1

\documentclass[11pt]{article}
\usepackage{graphicx}
\usepackage{tabularx}
\usepackage[]{acl}

\usepackage{times}
\usepackage{latexsym}
\usepackage{url}

\usepackage[T1]{fontenc}

\usepackage[utf8]{inputenc}

\usepackage{microtype}

\usepackage{inconsolata}

\newcommand{\furl}[1]{\footnote{\url{#1}}}

\title{Towards a Brazilian History Knowledge Graph}

\author{Valeria de Paiva \\
  Topos Institute, USA \\
  \texttt{valeria@topos.institute} \\\And
  Alexandre Rademaker \\
  FGV/EMAp, Brazil \\
  IBM Research, Brazil \\
  \texttt{alexandre.rademaker@fgv.br} \\}

\begin{document}
\maketitle

\begin{abstract}
This short paper describes the first steps in a project to construct a knowledge graph for Brazilian history based on the Brazilian Dictionary of Historical Biographies (DHBB) and Wikipedia/Wikidata. We contend that large repositories of Brazilian-named entities (people, places, organizations, and political events and movements) would be beneficial for extracting information from Portuguese texts. We show that many of the terms/entities described in the DHBB do not have corresponding concepts (or Q items) in Wikidata, the largest structured database of entities associated with Wikipedia. We describe previous work on extracting information from the DHBB and outline the steps to construct a Wikidata-based historical knowledge graph.
\end{abstract}

\section{Introduction}

Knowledge graphs (KG) are handy artifacts. As reported in several sources, automatic KG construction from text is a highly non-trivial and sought-after task suitable for many industrial applications.

This task is also essential for academic knowledge organization. But if this kind of construction is full of pitfalls in English, it is still more so in less-resourced languages, like Portuguese. In this short paper, we deploy recent tools of NLP, used to process natural language corpora, to start the development of a knowledge graph for Brazilian recent history, as described in the \textit{Dicionário Histórico Biográfico Brasileiro} (DHBB), in English the Brazilian Historical-Biographical Dictionary.

Section~\ref{sec:dhbb} describes the DHBB dictionary and how it has been maintained over the years and clearly states our goal. Section~\ref{sec:previous} reports some previous works applying NLP techniques to DHBB. Section~\ref{sec:wikis} discusses why DHBB is not merely ingested into Wikipedia but maintained as a self-contained project at the Getulio Vargas Foundation. In Section~\ref{sec:map} we presented our strategy to map the titles of DHBB articles to Wikidata entries, the most relevant issues and an evaluation of our mapping. Finally, we present our final considerations in Section~\ref{sec:conclusion}.

\section{The DHBB corpus}\label{sec:dhbb}

The Brazilian Historical-Biographical Dictionary (DHBB) was conceived to provide researchers and scholars of Brazilian History with organized and systematic information about personalities and themes considered noteworthy in the recent history of Brazil.

Although named a “dictionary,” the DHBB has an encyclopedic format, with long entries written by experts describing relevant actors and events in recent Brazilian history. The time frame covered by the DHBB encompasses the historical period that began with the ``Revolução de 1930'' (the 1930s Revolution), a period in Brazilian history marked by a rupture and significant renewal in the political elite. Currently, the DHBB has 7,863 entries covering people, institutions, organizations, and events. Most are biographical, with over 6,800 biographies and around a thousand thematic entries.

The project was initiated and is pursued by CPDOC (Centro de Pesquisa e Documentação de História Contemporânea do Brasil), an organization within the Fundação Getúlio Vargas (FGV), which has as its main goal the preservation of relevant document sources for the country's history and the development of research tools and methods for historical and economic research on the Brazilian cultural heritage. 

The first edition of the DHBB dates from 1984, and since then, some major updated editions have appeared in 2010, 2014, and 2015. The update in 2010 was innovative in that it brought full availability of the contents of the DHBB to the Internet, where it can be consulted at its online version~\furl{https://www.fgv.br/cpdoc/acervo/arquivo}. In 2014, a team of editors at CPDOC, in collaboration with researchers from the School of Applied Mathematics (EMAp) -- also a department of FGV -- was formed, and they managed to get the contents of the DHBB into a GitHub repository (\url{https://github.com/cpdoc/dhbb}). The DHBB is a benchmark of scholarly work, providing, in a concise and unified way, a significant amount of data that had been dispersed in several primary and secondary sources beforehand. The dictionary aims at objective and unbiased entries, avoiding ideological or personal judgments as much as possible. CPDOC researchers carefully revise all entries to ensure the information's accuracy and a uniform style across different entries.

The subsection of the DHBB concerned with \textit{thematic entries} is of particular interest to us in this work. The thematic entries of the DHBB describe political parties and political movements, organizations, and historical events. Also included are constitutions, decrees, laws, statutes, certain current and basic concepts of political history, economic and administrative institutions, national impact newspapers, and topics about foreign relations issues. One would like to guarantee that most of the precious information available in the DHBB can be utilized in modern NLP tools as structured data. So, we would like to ensure that all the named entities we can detect with Spacy (\url{https://spacy.io/}), for instance, are concepts present in Wikidata. If they are not, we should complete Wikidata to serve as a backbone for a Knowledge Graph for Brazilian History. However, the titles of the DHBB entries should be the most critical `named entities,' and it is expected that they should be present in Wikipedia/Wikidata already. 

Here, we are first connecting named entities in the DHBB corpus to Wikidata items. It was a surprise to realize that many could not be mapped automatically. Of course, there are many disambiguation problems to discuss.

\section{Previous work}\label{sec:previous}

Several computational linguistics research projects already use the DHBB as a primary source of information about Brazilian history. We can cite \citet{depaiva2014}, a very preliminary exploration of the dictionary using tools such as the syntactic analyzer FreeLing~\cite{padro:2012}. Freeling is an open-source, multilingual processing library providing a wide range of analysis functionalities. It was used with  OpenWN-PT~\cite{coling2012}, a Portuguese version of Princeton's WordNet database in that first exploration. 

Once the original text of the dictionary was made available in GitHub, the work in \citet{higuchi2018text} discusses (linguistic) \textit{appositives} and their use, creating lists to help extract the semantic information provided by them. This was followed by the exciting work of \citet{higuchi2019distant} on what they call `distant reading of History.' That work concerns what information one can extract from the DHBB corpus without reading all the entries individually. According to \citet{higuchi2019distant}, there are around 48,500 persons,  27,500 organizations, 5,000 places, and 36,900 other important named entities (like political movements and laws) in the DHBB. Given this vast material in the corpus, ours is a long-haul project.

The work by \citet{ribeiro2020construction} discusses different tools to detect errors in the syntactical processing of the DHBB corpus and looks for possible best methods to tackle the existing problems. In particular, this work discusses differences in segmentation of the text into sentences and tokenization between different lexical-syntactic frameworks. In a large corpus such as the DHBB, they found  312,539 sentences in the Linguateca~\footnote{Linguateca is a distributed language resource center for Computational Processing of Portuguese, see \url{https://www.linguateca.pt}.} processing, 314,930 sentences according to Apache's OpenNLP framework and 311,530 sentences according to Freeling.

More recently, work stemming from the `distant reading of History' of \citet{higuchi2019distant} has discussed the `age of entrance' in Brazilian politics, the `academic background of Brazilian politicians,' and family ties among the political elites in \citet{higuchi2022automatic}. These are fascinating questions for researchers interested in digital humanities. They corroborate the need for knowledge graphs for different cultures, as suggested by \citet{depaiva2022seringueiros}. This last work observes that  Portuguese terms specific to Brazil (e.g., typical fruits, fish, music, and occupations, amongst others) are less likely to appear in the already existing lexical resources. This is because many of the lexical resources built recently are multilingual, meaning they tend to be about entities that exist in many languages, not the ones specific to the Brazilian culture.

\section{Wikipedia vs. Wikidata}\label{sec:wikis}

The modern concept of a general-purpose, widely distributed, printed encyclopedia originated with Denis Diderot and the 18th-century French encyclopedists. Wikipedia, a free-content online encyclopedia written and maintained by a community of volunteers, was launched on January 15, 2001. In February 2022, the English edition of Wikipedia had grown to 6,740,267 articles. But different languages have very different sizes of Wikipedia, depending on their community of volunteers. As of November 2023, the Portuguese Wikipedia is the 18th largest Wikipedia by article count, containing 1,112,226 articles, less than one-sixth of the English Wikipedia. This should be contrasted to the number of Portuguese speakers in the world: Portuguese is (according to \furl{https://bit.ly/4b0GR9V}) the 8th most spoken language in the world in 2023. Portuguese has around 263 million native speakers, the most spoken language in South America.

We should note that articles in Wikipedia need to be \textbf{notable}; hence, perhaps not all Brazilian politicians and not all the five thousand Brazilian municipalities can be considered notable enough to have their own Wikipedia page. So, these politicians may not have Wikipedia pages, but in principle, they should all be part of Wikidata, which aspires to be as complete as possible. Thus, Wikidata could be the basic infrastructure for a Brazilian history and culture knowledge graph. To check how complete the information in Wikidata is about Brazilian modern history, we aim to map named entities in the DHBB corpus to concepts in Wikidata. We start this project by mapping the thematic and the biographical entries of the DHBB to Wikidata, using \textit{wikimapper}, a freely available library downloadable from \url{https://github.com/jcklie/wikimapper}.

It is worth remembering that the information in the DHBB is human-created and of high quality. Researchers at CPDOC might not want all their work reproduced on Wikipedia, and certainly, Wikipedia forbids whole copying of any work already in existence. But the entities that these works refer to, the names of historical characters, locations, organizations, and historical events (such as revolts, treaties, impactful laws, etc.) should be available in the structured form presented by Wikidata.

\section{Mapping Historical Brazilian Entities}\label{sec:map}

We first map the titles of the thematic entries of the DHBB to Wikidata. We know there are 973 such entries found at \url{https://github.com/cpdoc/dhbb-nlp/tree/master/wikidata}. However, we can only find Wikidata items for 498 of these 973 titles. So, even assuming that all the mappings are correct (which they are not), we only have 51\% of concepts for the thematic entries' titles in the DHBB.

Examples of thematic entries in the DHBB are presented in Table~\ref{tab:dhbbb}.

\begin{table}[htbp]
\small
\begin{tabularx}{\columnwidth}{lX}
id & title\\[0pt] \hline
\hline
11602 & Central Geral dos Trabalhadores do Brasil (CGTB)\\[0pt]
11603 & Conselho de Comunicação Social (CCS)\\[0pt]
11604 & Coordenação Nacional de Lutas (Conlutas)\\[0pt]
11605 & EXTRA\\[0pt]
11606 & Horário Gratuito de Propaganda Eleitoral (HGPE)\\[0pt]
11607 & Lei de Responsabilidade Fiscal\\[0pt]
11608 & Nova Central Sindical de Trabalhadores (NCST)\\[0pt]
11609 & Parcerias Público-Privadas (PPP)\\[0pt]
11610 & PARTIDO DA CAUSA OPERÁRIA (PCO)\\[0pt]
11611 & PARTIDO DA REPÚBLICA (PR)\\[0pt]
11612 & PARTIDO HUMANISTA DA SOLIDARIEDADE (PHS)\\[0pt]
11613 & PARTIDO PROGRESSISTA (PP)\\[0pt]
11614 & PARTIDO RENOVADOR TRABALHISTA BRASILEIRO (PRTB)\\[0pt]
11615 & PARTIDO REPUBLICANO BRASILEIRO (PRB)\\[0pt]
11616 & PARTIDO SOCIAL DEMOCRATA CRISTÃO (PSDC)\\[0pt]
11617 & PARTIDO SOCIAL LIBERAL (PSL)\\[0pt]
11618 & PARTIDO SOCIALISMO E LIBERDADE (PSOL)\\[0pt]
11619 & PARTIDO TRABALHISTA CRISTÃO (PTC)\\[0pt]
11620 & Pensamento Nacional das Bases Empresariais (PNBE)\\[0pt] \hline
\end{tabularx}
\caption{Examples of thematics entries in DHBB.}\label{tab:dhbbb}
\end{table}



We have not, yet, a reasonable mapping between DHBB titles of thematic entries and Wikidata concepts, but given that there are 973 titles, and we have 475 thematic titles that produce no Wikidata result whatsoever, we can conclude that a serious effort of completing the information of Wikidata with named entities from the DHBB is a necessary and useful project.

We start this project by producing a publicly available spreadsheet~\footnote{The data is being consolidated in the final repository at \url{https://github.com/cpdoc/dhbb-nlp/tree/master/wikidata} and in the metadata of the DHBB entries.} that historians and interested parties can consult and help us improve. Crowd-sourcing this disambiguation project is an exciting project for history professors.

Given that 498 thematic entries have some Q items in Wikidata, are these correct? There are some very wrong ones, e.g., political parties from other countries. Some entries are missed, even if the item is in Wikipedia and Wikidata.  For example, the infamous \textbf{DOI-CODI}  (full name ``Destacamento de Operações e Informações – Centro de Operações e Defesa Interna''), Brazil's intelligence and political repression agency during the military dictatorship (1964–1985) exists in Wikidata. But neither Wikimapper nor the Wikidata search API can find it automatically. This was manually found at \url{https://www.wikidata.org/wiki/Q5205864}.

We can only find some items like the \textit{Conselho de Comunicação Social (CCS)} using a web search engine. Perhaps because search engines find and index information from government sites, like the Chamber of Deputies of Brazil (\url{https://www.camara.leg.br/} ) has biographical information about all Federal Deputies since 1823. Clearly, Google's Knowledge Graph is more complete than Wikidata.

\subsection{Cleaning up issues}\label{sec:issues}

Given that the terms we are trying to map are organizations and events from Brazilian History, one can filter out political organizations in other countries. These sometimes have similar names, for example, the \textit{União Sindical Independente (USI)} in Brazil and \textit{União de Sindicatos Independentes} in Portugal. However, many international organizations do not have the property "country" in Wikidata.

A most frequent case is when we are trying to find a Wikidata or Wikipedia page for a concept that is part of a larger concept. Many times, only the larger concept has its own Wikipedia/Wikidata item. The smaller concept can be mentioned as a paragraph on the page for the more significant concept. For example, the organization \textit{Ação Democrática Popular} (\url{https://github.com/cpdoc/dhbb/blob/master/text/5705.text}) is a sub-item of the Wikipedia entry for the \textit{Instituto Brasileiro de Ação Democrática} (\url{https://bit.ly/48GPJjh}). This seems to be one of the most common cases of missing entries. Entities that no longer exist are also a significant part of the missed concepts. For example, the news magazine \textbf{Afinal}  was only published between 1984 and 1989 but had a DHBB thematic entry, \url{https://github.com/cpdoc/dhbb/blob/master/text/5724.text}. As far as we can see, this publication is not listed in either Wikipedia or Wikidata. 

Search engines index more text than simply names of entities, so using a search engine to find the Wikipedia article connected to a particular thematic entry in the DHBB usually leads to a more significant entity entry connected to the one you started from. This is not the case when a new event with the same entity intervenes. For instance, there was an `Agreement of Fernando de Noronha' (\url{https://github.com/cpdoc/dhbb/blob/master/text/5718.text}) in 1957 between Brazil and the US about using the island for launching missiles and also a `Brazil-US trade agreement' (\url{https://github.com/cpdoc/dhbb/blob/master/text/5717.text}). But recently, there was another disputed agreement between the state of Pernambuco and the  Brazilian Federal government about the jurisdiction over the Fernando de Noronha island. So, one can no longer find references to the old 1957 agreement, as most references are about the new agreement.

Some typos preclude the proper working of the search (e.g. the DHBB entry is called \textit{Ação Imperial Patrionovista} (\url{https://github.com/cpdoc/dhbb/blob/master/text/5707.text}) instead of the correct version for Wikipedia \textit{Ação Imperial Patrianovista} (\url{https://pt.wikipedia.org/wiki/Patrianovismo}), where again Google can find the reference, as it searches up to common misspellings. 

Overall, Wikidata does not have the necessary information about entities of Brazilian recent history, at least as far as the DHBB's thematic entries are concerned. Are the biographical entries better represented?

This would seem very unlikely, given the Wikipedia notoriety constraints and the difficulties of deciding when a  text refers to a specific person. There are several different people with the same name. Moreover, names in Portuguese are not as constrained as in English: people have several surnames and can use them differently. For example, the Brazilian diplomat `Paulo Tarso Flecha de Lima,' has the Wikipedia entry \url{https://en.wikipedia.org/wiki/Paulo_Tarso_Flecha_de_Lima}. However, he is called simply `Flecha de Lima' in his DHBB entry (\url{https://github.com/cpdoc/dhbb/blob/master/text/2865.text}) and so we cannot automatically find his DHBB entry. Nevertheless, out of 6980 biographic entries, we can find Wikidata items for 4300 names; hence, less than 38\% titles do not get a mapping at all. We still need to detect the right WikiData item, of course.

\subsection{Evaluation}\label{sec:eval}

Table~{tab:summary} summarizes the results of searching all DHBB entry titles (thematic+biographical) for related Q items in Wikidata, using Wikimapper. 

Given the situation described in the section above, with false positives and false negatives, we produce a human evaluation using a sample with 100 titles, 25 thematic ones that Wikimapper finds at least one Wikidata Q item, and 25 thematic titles without an automatically mapped Q item. Similarly, there are 25 with attached automatic Q items for biographical titles and 25 without Q items.

\begin{table}[hbtp]
  \begin{center}
    \begin{tabular}{lrr}
      & in Wikidata & not in Wikidata\\ \hline
      biographical & 4,300 & 2,590\\
      themed & 498 & 475\\ \hline
    \end{tabular}
  \end{center}
  \caption{The number of DHBB entries, by type, that using Wikimapper we could find in Wikidata or not.}\label{tab:summary}
\end{table}

Out of the 25 names of people that Wikimapper finds no Q items for, we realize that 30\%  can be found by a human. Also, for the 25 names of people (biographical entries) that Wikimapper can find entries, 16\% of the names automatically mapped are wrong, and we cannot find a correct mapping.

Similarly, 28\% of the thematic entries not found automatically can be found by a wilful human, while 34\% of the entries automatically mapped via Wikimapper are mistakes. Some are mapped 
to disambiguation pages in Wikidata, for instance. For prevalent names, for example, \textit{Francisco Silva}, we have 46 possible entries in Wikipedia/Wikidata, counting both Portuguese and Brazilian biographies. Still, of course, the DHBB refers to one specific person. But since this is found amongst the many `Franciscos,' we count it as a correct entry. 

It is unclear whether the proportions we found in this random sampling will hold for the whole task. We should increase the accuracy of our mappings, but we should also try to complete Wikidata with the terms genuinely not found.

\section{Conclusion}\label{sec:conclusion}

Building up knowledge graphs for academic subjects such as contemporary Brazilian History or modern Brazilian art seems vital to preserving our cultural heritage. Projects like the DHBB or the `Enciclop\'edia Cultural Ita\'u' (\url{https://enciclopedia.itaucultural.org.br/}) are precious. However, this material needs to be connected and structured so that we can pose queries and do some reasoning over it. Linking to Wikipedia/Wikidata should also help make the DHBB entries appear more in casual searches for the titles of the entries in search engines. Currently, the wealth of information on the DHBB is not being shared as widely as we would like to see it shared. Adding the kind of information in the DHBB to Wikidata seems a no-nonsense way of making this information more widely available. This is what our project is about.


\bibliography{custom}





\end{document}